# A Natural Law of Succession[1]

Eric Sven Ristad

**Abstract**

Consider the following problem. You are given an alphabet of $k$ distinct symbols and are told that the $i^{th}$ symbol occurred exactly $n_i$ times in the past. On the basis of this information alone, you must now estimate the conditional probability that the next symbol will be $i$. In this report, we present a new solution to this fundamental problem in statistics and demonstrate that our solution outperforms standard approaches, both in theory and in practice.

[1]This paper would be much the worse without the help of my colleagues at Princeton and LM'95. Thanks to Andrew Appel, Steve Golowich, Fred Jelinek, Hermann Ney, Andreas Stolcke, Robert Thomas, Kenji Yamanishi, and Andy Yao for their suggestions. Richard Jeffrey enriched my appreciation of prior work on inductive methods. George Miller and Matthew Norcross shared their knowledge of words and file systems, respectively. The author is partially supported by NSF Young Investigator Award IRI-0258517.

# 1 Introduction

Consider the following problem. You are given an alphabet $A$ of $k$ distinct symbols and are told that the $i^{th}$ symbol occurred exactly $n_i$ times in the past. On the basis of this information alone, you must now estimate the conditional probability $p(i|\{n_i\}, n)$ that the next symbol will be $i$. We present a new solution to this fundamental problem in statistics and inductive reasoning and demonstrate that our solution outperforms standard approaches, both in theory and in practice.

This problem of multinomial estimation is arguably the single most important parameter estimation problem in all of statistics. Its solution is necessary in any application requiring a discrete probability function, such as classification with decision trees or time series prediction.

For time series prediction with Markov processes, the alphabet $A$ is the alphabet of the Markov process and the $\{n_i\}_s$ are the number of times that the $i^{th}$ symbol has been generated in the state $s$. The prediction $p(i|\{n_i\}_s, n_s, s)$ represents the probability that the state $s$ will generate the $i^{th}$ symbol and transition to the next state $\delta(s,i)$. As an example, consider the problem of predicting the next word in an English utterance based on the preceding words. The alphabet $A$ represents all the possible words of English and the states of the Markov process represent the immediately preceding words.

For classification with decision trees, the alphabet $A$ is the set of classes and the $\{n_i\}_s$ are the number of times that the $i^{th}$ class has been observed in a given equivalence set $s$. The equivalence sets correspond to the leaves of the decision tree. They partition the universe of objects based on their observable properties (ie., their features) while the classes $A$ represent a hidden property. The prediction $p(i|\{n_i\}_s, n_s, s)$ represents the probability that the next object that belongs to the equivalence set $s$ will belong to the $i^{th}$ class. As an example, consider the problem of medical diagnosis. The classes $A$ represent all the possible illnesses that a patient might have and the equivalence sets represent all the logically possible combinations of symptoms that a patient might have.

We begin by presenting the original solution to this problem — Laplace's law of succession — as well as the most widely-adopted practical solution — Lidstone's law of succession. Next we present our natural law of succession and compare it to the two standard laws. We show that the penalty for using either of the standard laws instead of the natural law grows without bound for any finite string drawn from a subset of the alphabet. Unlike prior analyses, our analysis applies to any finite string without reference to hidden source probabilities or to convergence in the limit. We conclude by comparing the empirical performance of the natural law to four parameter estimation rules employed by the data compression community. The natural law gives the best overall compression, by a wide margin. In an appendix, we compare the natural law to two variants of the classic Good-Turing estimate [12].

## 2   Parameter Estimation

The problem of predicting symbols from their observed frequencies is traditionally viewed as a problem in parameter estimation. Under this view, each symbol $i$ is associated with a hidden parameter $p_i$ whose value represents its probability of occurrence. Our task is to determine the value of all the parameters, using the observed frequencies. Let us consider two approaches to this parameter estimation problem, one due to Laplace and the other to Lidstone.

### 2.1   Laplace's Law

The original solution to this parameter estimation problem is Laplace's law of succession (1775)

$$p_L(i|\{n_i\}, n) = (n_i + 1)/(n + k) \qquad (1)$$

where $n = \sum_i n_i$ is the total number of symbols observed. Laplace proved that (1) is the correct Bayesian estimate when starting from a uniform prior on the symbol probabilities.

Laplace's law of succession has been widely criticized for assigning both too much and too little probability to novel events. For example, consider the question of whether or not the sun will rise tomorrow, given that it has risen $n_1$ times in the past and failed to rise $n_0$ times. In Laplace's circumstances, all the reports of history for the past five thousand years indicated that the sun had never failed to rise. So $n_0 = 0$ days and $n_1 = 1,826,213$ days, and therefore Laplace calculates the probability that sun will rise tomorrow to be $1 - 1/1,826,215$. For comparison, consider the popular New Jersey "Pick 6" state lottery. In order to win, one must pick the correct six balls from a set of 46, without replacement. There are 9,366,819 ways to do this, so the chance that the sun would fail to rise tomorrow is more than five times greater than the chance of winning Pick 6! Given a similar payoff, the inhabitants of New Jersey who subscribe to Laplace's law should be willing to bet on the sun's failing to rise for at least another 20,425 years.

Or consider the question of how many elephants live on the planet Mars. Mars has a surface area of approximately 22,166,424,000 square miles. Let us say for the sake of argument that each square mile can support at most one elephant, allowing for grazing and a comfortable life style. Then *a priori* we know that there are anywhere from 0 to 22,166,424,000 elephants on Mars. Having no other prior knowledge, we have no rational basis to distinguish between these possibilities and therefore assign them a uniform prior. Due to the frivolous nature of our project, we are only permitted to make a single interplanetary observation of Mars per day. Using Laplace's law of succession it would take 110,829,935,214 *millenia* in order for us to believe that there are no elephants on Mars with the same confidence that Laplace believed that the sun would rise tomorrow.

Finally, consider the problem of predicting the output of a random number generator that purports to generate 32 bit numbers uniformly. After $2^{32} - 1$ trials, you observe that this alleged random number generator has generated exactly $2^{32}-1$ distinct numbers! Only a single number $i$ has not been generated. According to Laplace's law of succession, the probability that the next number is $i$ is only $1/(2^{33} - 1)$.

The widely proclaimed absurdity of these results has led many influential statisticians to reject Bayesian inference altogether [8, 9, 19, 38]. Even the staunchest advocates of Bayesian reasoning feel compelled to give stern warnings against the use of Laplace's law [11, 13, 16, 17]. Apparently the risk of using Laplace's law has become so great that it may no longer be taught to undergraduates, even at MIT [1, 7].

It is only fair to point out, as Laplace did, that these calculations did not include our comprehensive prior knowledge of celestial mechanics, zoology, Martian meteorology, or computer programs.[2] But we may well wonder — what is the value of inductive inference if prior knowledge is its certain prerequisite?

## 2.2 Lidstone's Law

In statistical practice, a commonly adopted solution to the problem of multinomial estimation is Lidstone's law of succession for some positive $\lambda$.

$$p_\lambda(i|\{n_i\}, n) = (n_i + \lambda)/(n + k\lambda) \qquad (2)$$

This class of probability estimates is due to the actuaries G.F. Hardy [14] and G.J. Lidstone [28] at the turn of the century. These estimates were later explained by W.E. Johnson [21] as a linear interpolation of the maximum likelihood estimate $n_i/n$ and the uniform prior $1/k$. This may immediately be seen by rewriting (2) as (3) with the substitution $\mu = n/(n + k\lambda)$ [13].

$$p_\mu(i|\{n_i\}, n) = \mu \frac{n_i}{n} + (1 - \mu)\frac{1}{k} \qquad (3)$$

Johnson's perspective was taken up by R. Carnap in his widely-cited work on inductive reasoning [3].

This class of probability estimates is also prominent in the information theory literature [6, 10, 24, 25] and the statistical language modeling community [20]. The most widely advocated single value for $\lambda$ is $1/2$, for a diverse set of reasons [18, 31, 26]. For this reason, the special case $\lambda = 1/2$ has even been given its own name, namely, the Jeffreys-Perks law of succession [13]. Empirical investigation shows that the optimal value for $\lambda$ in data compression applications ranges from 1 for object code files to $1/32$ for files containing English prose.

---

[2]Immediately after calculating the probability of the sun rising, Laplace notes that "this number would be incomparably greater for one who, perceiving . . . the principle regulating days and seasons, sees that nothing at the present moment can check the sun's course." [27, p.11]

The value of the flattening parameter $\lambda$ represents the statistician's prior knowledge. It must be chosen before making any observations. Consequently, we are confronted with the dilemma of choosing $\lambda$ solely on the basis of our prior knowledge. In most cases, the necessity of choosing a parameter value *a priori* defeats the purpose of inductive inference. If we are required to study celestial mechanics in order to set the value of $\lambda$ for accurate sunrise prediction, then we may as well just ignore the observations altogether and simply predict that the sun will certainly rise tomorrow. But what then will we do when our prior knowledge is incomplete or not easily quantifiable? In such situations, we are conceptually obligated to set $\lambda$ to unity and bear the consequences of Laplace's law of succession.

In point of fact, no value of $\lambda$ is appropriate for sunrise prediction. $\lambda < 1$ represents greater trust in the empirical probabilities while $\lambda > 1$ represents less trust in the empirical probabilities. In the case of sunrise prediction, our confidence that the sun will rise tomorrow is based on our scientific understanding of celestial mechanics and the solar system, not on the mere fact that the sun has risen for the past 5,000 years. This line of reasoning suggests an enormous value for $\lambda$, which yields poor predictions. Consequently, no value of $\lambda$ is appropriate for this situation, although smaller $\lambda$ will give better predictions.

Nor does Lidstone's law offer a practical solution to the Martian elephants problem. To a first approximation, halving $\lambda$ only halves the number of observations required to achieve a given level of confidence. Since the set of symbols is so large, $\lambda$ must be essentially zero in order for the probability estimate to converge in our lifetimes. But then a single faulty observation, or the prank of an undergraduate astronomy student, will suffice to convince us of the existence of Martian elephants.

Finally, we note that only extremely large values of $\lambda$ will give better predictions for the defective random number generator. Unfortunately, a large $\lambda$ represents prior confidence in the uniformity of the random number generator. Just as a small $\lambda$ does not represent a lack of prior knowledge, a large $\lambda$ does not represent a wealth of prior knowledge. And thus we see that $\lambda$ does truly not represent our prior knowledge — rather, it only represents the prior knowledge that we wished we had after making our observations.

## 3    Natural Strings

Let us now consider a different approach to this problem. Instead of estimating parameter values, let us impose constraints on strings so that simple strings are more probable than complex ones. Our approach is principally inspired by the theory of stochastic complexity [34, 35, 36] and the related theory of algorithmic complexity [4, 23, 37].

The most important constraint on a string arises from its definition. A string is, by definition, a sequence of discrete symbols drawn from a finite alphabet with replacement. A uniform prior on all strings results in the probability assignment (4) for the string $x^n$ of length $n$ over an alphabet $A$ of size $k$.

$$p_L(x^n|n) = \left( \binom{n+k-1}{k-1} \binom{n}{\{n_i\}} \right)^{-1} \qquad (4)$$

This probability assignment corresponds to a uniform prior on all possible partitions of $n$ into $k$ subsets, followed by a uniform prior on all distinguishable strings that contain exactly $n_i$ occurrences of the $i^{th}$ symbol. It entails Laplace's law of succession, regardless of whether we define $p_L(i|x^n, n)$ as the conditional probability $p_L(x^n i|n+1)/p_L(x^n|n)$ or as the relative odds $p_L(x^n i|n+1)/\sum_{j \in A} p_L(x^n j|n+1)$.

The next most important property of a string — and the single most important property of a natural string — is that it is drawn from a proper subset of the alphabet. This is necessarily so for extremely short strings, but it is also true for long strings. For example, computer files are strings over an alphabet of size 256. Some files, such as those containing object code or executables, use all 256 symbols. Most files, however, are limited to a subset of the 95 printing and spacing ASCII characters. The entire King James bible contains only 66 distinct characters. Or consider that most English speakers use less than 20,000 of the more than 1,000,000 word forms listed in Webster's 3rd. Even worse, the set of word forms known by English speakers — of which Webster's 3rd is only a proper subset — is continually growing. No natural English sentence could possibly contain all the English word forms!

We consider two interpretations of the constraint that natural strings are drawn from a subset of the symbols in their alphabets. In the first interpretation, all nonempty subsets of the alphabet are equally likely. In the second interpretation, all nonzero subset cardinalities are equally likely.

If we assign a uniform prior to all nonempty subsets of the alphabet, then we obtain the following probability assignment,

$$p_S(x^n|n) = \left( \left( \sum_{i=1}^{\min(k,n)} \binom{k}{i} \right) \binom{n-1}{q-1} \binom{n}{\{n_i\}} \right)^{-1} \qquad (5)$$

where $q$ is the number of distinct symbols in $x^n$, ie., $q = |\{i : n_i > 0\}|$. The first term represents the number of nonempty subsets containing less than or equal to $\min(k, n)$ symbols. Any subset of size greater than $\min(k, n)$ is impossible; when $n \geq k$, there are exactly $2^k - 1$ nonempty subsets of $A$. The second term represents a uniform prior on the $n_i$, given that $k - q$ of the $n_i$ must be zero and the remaining $q$ must be greater than zero. The third term represents a uniform prior on all strings that contain exactly $n_i$ occurrences of the $i^{th}$ symbol, for each $i$. Since the vast majority of the subsets are of moderate size, both very large subsets and very small subsets are relatively improbable under (5).

If we assign a uniform prior to all nonzero subset cardinalities, then we obtain the following probability assignment instead.

$$p_C(x^n|n) = \left( \min(k,n) \begin{pmatrix} k \\ q \end{pmatrix} \begin{pmatrix} n-1 \\ q-1 \end{pmatrix} \begin{pmatrix} n \\ \{n_i\} \end{pmatrix} \right)^{-1} \quad (6)$$

The first term of the product represents a uniform prior on the number $q$ of attested symbols, $1 \leq q \leq \min(k,n)$. The second term represents a uniform prior on the subsets of $A$ that are of size $q$. The third term represents a uniform prior on the $n_i$, given that $k - q$ of the $n_i$ must be zero and the remaining $q$ must be greater than zero. The fourth term represents a uniform prior on all strings that contain exactly $n_i$ occurrences of the $i^{th}$ symbol, for each $i$.

In comparison to the uniform subsets prior (5), the uniform cardinality prior (6) assigns more probability to small and to large subsets of the alphabet, and less to subsets of moderate cardinality. For example, the uniform cardinality prior assigns higher probability to subsets of cardinality less than 110 or greater than 146 in an alphabet size of 256. Most computer files contain less than 95 characters or all 256, so the uniform cardinality prior will have an inherent advantage over the uniform subsets prior in the domain of computer files. Our investigation in section 5 confirms this prediction.

Neither probability function (5, 6) is Kolmogorov compatible because $p(i^n|n) = p(i^{n+1}|n+1)$ for every symbol $i$ in $A$. Every $A^n$ has a different model. Consequently, the probability assigned to a string is *not* the product of the conditional probabilities of every symbol in the string, ie., $p(x^n|n) \neq \prod_{t=0}^{n-1} p(x_{t+1}|x^t, t)$. In order to obtain a law of succession, we must therefore calculate the conditional probability $p(i|x^n, n)$ as the relative odds that the symbol $i$ will succeed $x^n$ relative to the probability than any symbol $j$ in $A$ will succeed $x^n$.

$$p(i|x^n, n) \doteq \frac{p(x^n i|n+1)}{\sum_{j=1}^{k} p(x^n j|n+1)} \quad (7)$$

Since our probability assignments depend only on the symbol frequencies $\{n_j\}$ in the string, $p(i|\{n_j\}, n) = p(i|x^n, n)$.

Algebraic manipulation on the uniform subsets prior (5) gives us the following law of succession (8), which is valid even when $n < k$.

$$p_S(i|\{n_i\}, n) = \begin{cases} (n_i + 1)(n + 1 - q)/((n+q)(n+1-q) + q(k-q)) & \text{if } n_i > 0 \\ q/((n+q)(n+1-q) + q(k-q)) & \text{otherwise} \end{cases} \quad (8)$$

Note that $p_S(i|\{n_i\}, n) = (n_i + 1)/(n + k)$ if $q = k$, and so the uniform subsets law reduces to Laplace's law when all the symbols are attested.

Similarly, the uniform cardinality prior (6) yields the following law of succession (9), which we call the natural law.

$$p_C(i|\{n_i\}, n) = \begin{cases} (n_i + 1)/(n + k) & \text{if } q = k \\ (n_i + 1)(n + 1 - q)/(n^2 + n + 2q) & \text{if } q < k \land n_i > 0 \\ q(q+1)/(k-q)(n^2 + n + 2q) & \text{otherwise} \end{cases} \quad (9)$$

A central difference among the four laws of succession is the amount of probability they assign to novel symbols. By a slight abuse of notation, let $p(q|\{n_i\}, n) \doteq \sum_{\{i:n_i>0\}} p(i|\{n_i\}, n)$ be the total probability assigned to attested symbols and let $p(\bar{q}|\{n_i\}, n) \doteq 1 - p(q|\{n_i\}, n)$ be the total probability assigned to novel symbols.[3]

$$\begin{array}{rcl}
p_L(\bar{q}|\{n_i\}, n) & = & (k-q)/(n+k) \\
p_\lambda(\bar{q}|\{n_i\}, n) & = & (k-q)\lambda/(n+k\lambda) \\
p_S(\bar{q}|\{n_i\}, n) & = & q(k-q)/((n+q)(n+1-q) + q(k-q)) \\
p_C(\bar{q}|\{n_i\}, n) & = & q(q+1)/(n^2 + n + 2q)
\end{array} \qquad (10)$$

Now the difference between the estimates is striking. Unlike the standard laws of succession (1,2), the amount of probability assigned to novel events by our laws of succession (8,9) decreases quadratically in the number $n$ of trials. Doubling the number of trials will decrease the likelihood of novel events by a factor of four. Even more importantly, unlike the other three laws, the amount of probability assigned to novel events by the uniform cardinality law (9) is independent of the alphabet size $k$. There is no penalty for large alphabets.

The direct result is that the uniform cardinality law makes extremely strong predictions. For example, it takes only 1,911 days for the uniform cardinality law to achieve the same confidence that the sun will rise that Laplace's law achieved in 1,826,213 days. If we spent the same paltry 1,911 days counting the number of elephants on Mars, we would have achieved the same confidence that there are no Martian elephants that required Laplace's law 110,829,935,214 *millenia*!

Unlike Lidstone's law with $\lambda < 1$, the uniform cardinality law does not merely assign less probability to novel events. The amount of probability assigned to novel events depends on $q$ and on $n$. Recall our defective random number generator that emitted $2^{32} - 1$ distinct numbers in $2^{32} - 1$ trials. At that point, Lidstone's law would predict the lone remaining novel symbol with probability $\lambda/((1+\lambda)2^{32} - 1)$, ie., as a practical impossibility. In contrast, the uniform cardinality law would predict that novel symbol with probability $2^{32}/(2^{32} + 1)$, ie., as a virtual certainty.

Thus, the uniform cardinality law overcomes the practical difficulties encountered by Laplace's law without any prior knowledge of celestial mechanics, zoology, Martian meteorology, or computer programming.

The uniform cardinality law works because it wastes so little probability. It assigns such a low conditional probability to novel symbols only because it has assigned such a high probability to the previous symbols. In contrast, the standard laws of succession assign a much larger conditional probability to novel symbols, but only because they have assigned such a low probability to the previous symbols.

---

[3]The latter probability, $p(\bar{q}|\{n_i\}, n)$, is the "escape probability" of the data compression community [2] and the "backoff probability" of the speech recognition community [22]. See our discussion in section 5 below.

To see this, let's reconsider Laplace's famous sunrise example. According to Laplace's law, the total probability $p_L(\{n_1 = n+1\}|n+1)$ that the sun will rise for $n+1$ consecutive days is $1/(n+2)$, while the total probability $p_L(\{n_0 = 1, n_1 = n\}|n+1)$ that the sun will rise for all but one of the $n+1$ consecutive days is also $1/(n+2)$. According to the uniform cardinality law, the total probability $p_C(\{n_1 = n+1\}|n+1)$ that the sun will rise for $n+1$ consecutive days is $1/4$, while the total probability $p_C(\{n_0 = 1, n_1 = n\}|n+1)$ that the sun will rise for all but one of $n+1$ days is $1/2n$. Thus, $p_L(\{n_0 = 1, n_1 = n\}|n+1)$ is only about twice as likely as $p_C(\{n_0 = 1, n_1 = n\}|n+1)$. The tremendous disparity between Laplace's law and the uniform cardinality law arises because $p_C(\{n_1 = n+1\}|n+1)$ is about $n/4$ times as likely as $p_L(\{n_1 = n+1\}|n+1)$! Recall from (7) above that

$$p_C(1|1^n, n) \doteq \frac{p_C(1^n 1|n+1)}{p_C(1^n 1|n+1) + p_C(1^n 0|n+1)}$$

The enormous conditional probability $p_C(1|1^n, n)$ in favor of the sun continuing to rise is more due to the surprisingly high probability of $p_C(1^n 1|n+1)$ than to the low probability of $p_C(1^n 0|n+1)$. So our law of succession should satisfy those who believe that Laplace's law assigns too much probability to novel events as well as those who believe that it doesn't assign enough probability to novel events.

In statistical practice, it may prove desirable to reduce the flattening in our two laws of succession (8,9). The resulting "sharpened" estimates (11,12) more quickly approach the maximum likelihood estimate $n_i/n$ while still reserving the proper amount of probability for novel events.

$$p_{S'}(i|\{n_i\}, n) = \begin{cases} \left(\frac{n_i}{n}\right) \frac{(n+q)(n+1-q)}{(n+q)(n+1-q) + q(k-q)} & \text{if } n_i > 0 \\ \frac{q}{(n+q)(n+1-q) + q(k-q)} & \text{otherwise} \end{cases} \quad (11)$$

$$p_{C'}(i|\{n_i\}, n) = \begin{cases} \frac{n_i}{n} & \text{if } q = k \\ \left(\frac{n_i}{n}\right) \frac{n(n+1) + q(1-q)}{n^2 + n + 2q} & \text{if } q < k \wedge n_i > 0 \\ \left(\frac{1}{k-q}\right) \frac{q(q+1)}{n^2 + n + 2q} & \text{otherwise} \end{cases} \quad (12)$$

These laws give better results when estimating the state transition probabilities of a large Markov model, or in any other situation where the total number of trials is small.

## 4  Analysis

The practical importance of strong predictions confined to subsets of the alphabet cannot be overstated. Let us therefore determine which law gives the best predictions when strings are confined to a subset of the alphabet. We propose two analyses to answer this question. The first analysis reports the total probability assigned to the set of strings generated from a subset of the alphabet. We show that the total probability assigned to the set of possible strings by Lidstone's law rapidly approaches zero. In contrast, the total probability assigned to the possible string by the natural law is a constant, independent of the string length. The second analysis considers the ratio of the probabilities assigned by two laws of succession to any string drawn from a subset of the alphabet. We show that the natural law assigns more probability to any such string than is assigned by either of the standard laws. Both of our analyses have the singular virtue of applying to finite strings, without reference to hidden source probabilities.

Let us first consider the case where the set of possible strings is drawn from a subset of the alphabet. Let $A$ be the universe of possible symbols and let $B$ be the actual alphabet of the source, $B \subset A$, and $|B| = b$. In our examples, $A$ might be the set of all 256 8-bit bytes and $B$ might be the set of 95 printing and spacing ASCII characters. One important question we can ask of a probability estimate is how much probability it wastes on the impossible strings $A^n - B^n$ of a given length $n$, without prior knowledge of $B$.[4]

According to Laplace's law, there are $(n+k-1)!/n!(k-1)!$ logically possible $\{n_i\}$ although in the current situation only $(n+b-1)!/n!(b-1)!$ are in fact possible for our source. Therefore

$$
\begin{aligned}
p_L(B^n|n) &= \binom{n+b-1}{b-1} / \binom{n+k-1}{k-1} \\
&= \frac{(k-1)!}{(b-1)!} \frac{(n+b-1)!}{(n+k-1)!} \\
&< \left(\frac{k-1}{n+k-1}\right)^{k-b}
\end{aligned}
$$

and the total probability $p_L(B^n|n)$ assigned to the possible strings rapidly approaches zero for increasing $n$ when $b < k$. Equivalently, the total probability $p_L(A^n - B^n) = 1 - p_L(B^n)$ wasted on the impossible strings approaches unity when $b < k$.

A similar result obtains for Lidstone's law. The total probability assigned

---

[4]In the most general case, we might consider the amount of probability assigned by an estimate to an arbitrary set $\mathcal{B}_n$ of all possible strings, $\mathcal{B}_n \subseteq A^n$, without prior knowledge of the impossible strings $\overline{\mathcal{B}}_n$. Such an analysis reveals how quickly the estimate distinguishes the possible from the impossible, and how much probability is ultimately wasted on the impossible.

to a string $x^n$, $x^n \in A^n$, with symbol frequencies $\{n_i\}$ by Lidstone's law is

$$p_\lambda(x^n|n) = \frac{\prod_i(\Gamma(n_i + \lambda)/\Gamma(\lambda))}{\Gamma(n + k\lambda)/\Gamma(k\lambda)} \tag{13}$$

while the total probability assigned to the possible strings $B^n$ is as follows, using Stirling's approximation.

$$\begin{aligned}
p_\lambda(B^n|n) &= \prod_{i=0}^{n-1} \frac{i + b\lambda}{i + k\lambda} \\
&\approx \frac{\Gamma(k\lambda)}{\Gamma(b\lambda)} \left(\frac{1}{n-1}\right)^{\lambda(k-b)} \\
&= \Theta\left(\left(\frac{1}{n-1}\right)^{\lambda(k-b)}\right)
\end{aligned}$$

Unless $\lambda = 0$ and $b = 1$, the total probability $p_\lambda(B^n|n)$ assigned to the possible strings approaches zero as $n$ approaches infinity.

Thus, Lidstone's law without prior knowledge of $B$ performs arbitrarily worse than any other probability function with prior knowledge of $B$. This profound flaw in Lidstone's law — and in Laplace's law — is typically disguised by analyzing how the probability estimate $p_\lambda(i|\{n_i\}, n)$ converges to the underlying source probability as $n$ approaches infinity. Our much simpler analysis shows that convergence of probabilities in the limit is at best a feeble optimality.

How much probability does the uniform cardinality law waste on the impossible strings? According to the uniform cardinality law, there are $kk!(n-1)!/q!(k-q)!(n-q)!(q-1)!$ logically possible $\{n_i\}$. To simplify the analysis, let us consider the case where $n$ is sufficiently large so that $q = b$. (The case $q < b$ is more favorable to the uniform cardinality law.) Then only $(n-1)!/(n-b)!(b-1)!$ of the $\{n_i\}$ are still possible, and

$$\begin{aligned}
p_C(B^n) &= \binom{n-1}{b-1} \bigg/ k\binom{k}{b}\binom{n-1}{b-1} \\
&= \frac{b!(k-b)!}{(k+1)!}
\end{aligned}$$

which is minimized for $b = k/2$. At that point, the total probability $p_C(B^n)$ assigned to the possible strings is strictly greater than the constant $2^{-k/2}$ irrespective of $n$. Thus, the uniform cardinality law without prior knowledge of $B$ performs at most a constant factor worse than *any* other probability function with prior knowledge of $B$!

As our second theoretical question, let us consider how much the probabilities assigned by the uniform cardinality law can differ from those given by Laplace's law. In particular, how can we bound the ratio $p_C(x^n|n)/p_L(x^n|n)$ for all $n$ and $x^n$? Observe that the ratio of the two probabilities depends only on $n$, $k$, and $q$. Without loss of generality, we assume $n \geq k$ to simplify the analysis. (The case $n < k$ is considerably more favorable to the uniform cardinality law.)

$$p_C(x^n|n)/p_L(x^n|n) = \binom{n+k-1}{k-1} / k \binom{k}{q} \binom{n-1}{q-1}$$
$$= \Theta((n+k-1)^{k-q})$$

In other words, encoding a sequence of length $n$ with Laplace's law instead of the uniform cardinality law will cost an additional $\Theta((k-q)\log n)$ bits. The penalty for using Laplace's law instead of the uniform cardinality law grows without bound.

A similar result obtains for Lidstone's law.

$$p_C(x^n|n)/p_\lambda(x^n|n) = \Theta(\lambda^{(1-k)/2} n^{k\lambda-q} \prod_{i=1}^{k} n_i^{1-\lambda})$$

We include the $\lambda^{-1}$ term as a reminder that $\lambda = 0$ will result in an infinite advantage for the uniform cardinality rule when $q > 1$. This ratio is a function of the parameter $\lambda$ and the observed $\{n_i\}$. For the Jeffreys-Perks law of succession, with the widely advocated $\lambda = 1/2$, the penalty grows without bound when $q \leq k/2$. The case $q > k/2$ depends on the actual symbol frequencies. When the entropy of the relative frequencies is high and the number of attested symbols is high, then the uniform cardinality law will provide better predictions than the Jeffreys-Perks law. When the empirical entropy is low but the number of attested symbols is large, then the Jeffreys-Perks law will provide better predictions than the uniform cardinality law.

The predictions given by the uniform subsets law differ from those of the uniform cardinality law by a constant that depends only on $k$ and $q$, independent of $n$. Consequently, the uniform subsets law also enjoys a significant advantage over Laplace's law and Lidstone's law.

Finally, we note that our analysis applies equally well to finite memory models whose state transition probabilities are estimated using either Laplace's law [32, 33] or the Jeffreys-Perks law [39, 40]. Unless all the alphabet symbols are observed in all the model states – an incredibly unnatural situation – the total probability assigned to the possible strings by such models rapidly approaches zero as the strings get longer. Estimating the state transition probabilities of finite memory models using the natural law instead of the Jeffreys-Perks law will result in a prediction advantage that grows as a polynomial in the length of the string, even when the string includes all the symbols of the alphabet.

# 5 Application

Data compression provides a stringent empirical test for all probability estimates. Here the task is to predict the next byte in a file on the basis of the (frequencies of the) preceding bytes. Better probability estimates provide greater compression. So let us compare our two laws of succession (8,9) to the best estimates that the data compression community has to offer.

The four most widely used probability estimates in the data compression community, affectionately dubbed "methods A–D," are summarized in the following table

| Method | $p(i\|\{n_i\}, n)$ | $p(\bar{q}\|\{n_i\}, n)$ |
|---|---|---|
| A [5] | $n_i/(n+1)$ | $1/(n+1)$ |
| B [5] | $(n_i - 1)/n$ | $(k-q)q/(k-q')n$ |
| C [29] | $n_i/(n+q)$ | $q/(n+q)$ |
| D [15] | $(n_i - \frac{1}{2})/n$ | $q/2n$ |

where $q' \doteq |\{i : n_i > 1\}|$ is the number of symbols that have occurred at least twice. The second column gives the conditional probability of attested symbols while the third column gives the total probability assigned to novel symbols (the so-called "escape probability" for context models).[5]

The sole justification for these estimates is their empirical performance in data compression applications. In fact, the authors of methods A–B go so far as to claim that "there can be no theoretical justification for choosing any particular escape [probability] as the optimal one." [2, p.145] Fortunately, our work provides a clear theoretical justification: pick the method that most closely approximates the uniform cardinality law. In most cases, $n \gg q$, and then method D most closely approximates the uniform cardinality law. Method B provides the worst approximation to the uniform cardinality law because it assigns such low probability to symbols that have occurred exactly once. Consequently, our approach predicts that out of the four methods A–D, B is the worst and D is the best. Let the tests begin.

Our first test is sunrise prediction. Since we believe that the sun has risen for the past 5,220 years of recorded history, the test file contains 1,906,605 bytes, all '1' to indicate a successful sunrise.[6] Since $q = 1$, methods A and C give the same predictions in this situation. The slope of the compression curve represents the rate at which the estimation rule becomes confident that the sun will rise tomorrow, where a flatter compression curve represents stronger confidence in tomorrow's sunrise. Results are graphed in figures 1, 2.[7]

---

[5] Method B assigns probability $(n_i - 1)/n$ to symbols that have occurred at least twice, and then assigns the remaining $q/n$ probability uniformly to all symbols that have occurred less than two times [5]. There are exactly $k - q$ novel symbols, so the total probability assigned to novel symbols by method B is $((k - q)/(k - q'))(q/n)$.

[6] The reader will note that this task is more difficult than Laplace's sunrise prediction problem because the alphabet size is 256 instead of 2. Those that find this objectionable may imagine the task of sunrise prediction to be to predict whether the sun will rise, and if not, then why not. A list of 255 possible reasons why the sun might fail to rise is available from the author upon request.

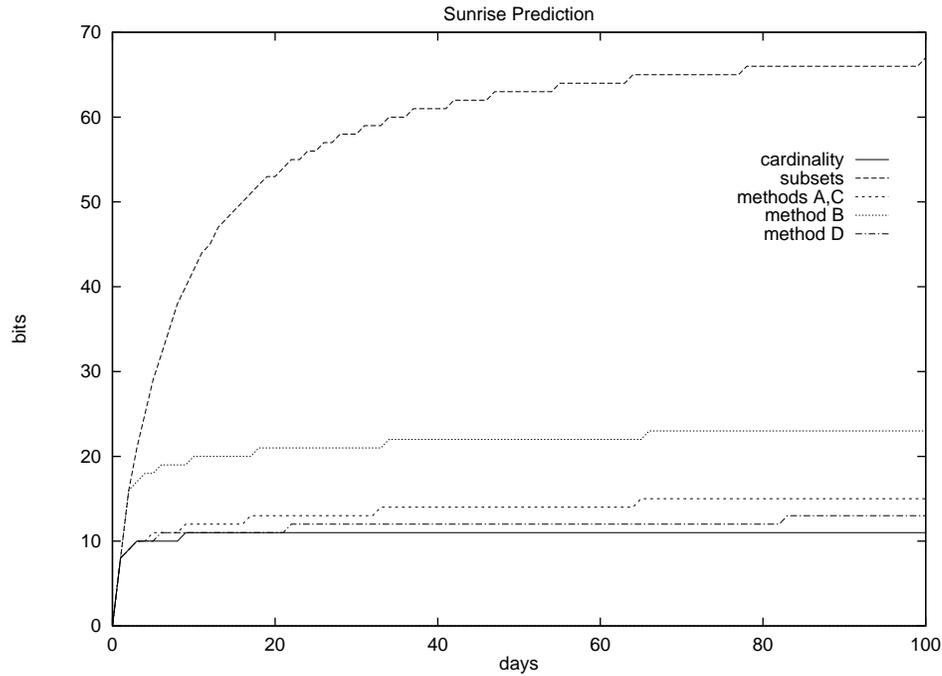

Figure 1: The first 100 days of recorded history. The horizontal axis represents days while the vertical axis represents the negative log of the total probability that the sun will rise for $n$ consecutive days. A flatter curve represents greater confidence in tomorrow's sunrise.

---

[7] Neither Laplace's law nor the Jeffreys-Perks law are graphed because their performance on this file is so bad. Laplace's law encodes the first 100 days in 301 bits and the entire 5,220 years in 3,644 bits. The Jeffreys-Perks law encodes first 100 days in 225 bits and the entire 5,220 years in 1,952 bits.

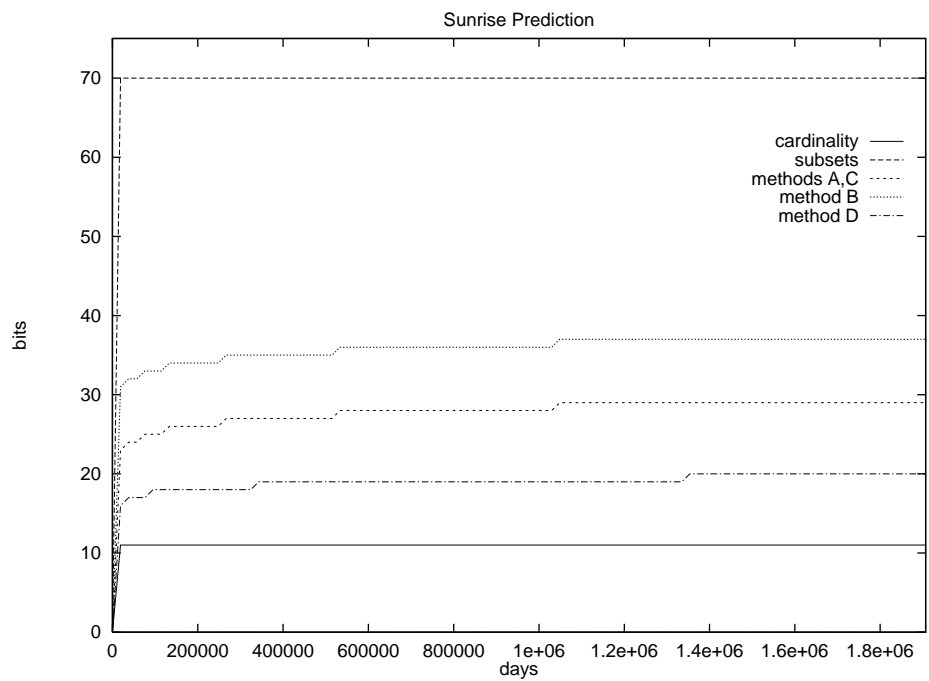

Figure 2: The past 5,220 years of recorded history. The horizontal axis represents days while the vertical axis represents the negative log of the total probability that the sun will rise for $n$ consecutive days (in bits). A flatter curve represents greater confidence in tomorrow's sunrise. After only 9 days, the uniform cardinality law is extremely confident in the sunrise. The uniform subsets law becomes equally confident after 530 days. In contrast, methods A-D do not achieve the same level of confidence any time during the 5,220 years of recorded history.

Our second test is the Calgary data compression corpus, which includes a wide range of ASCII as well as non-ASCII files [2].[8] Our compression results are summarized in the following table 1. All compression results are in whole bytes, rounded up. Again, the uniform cardinality law is the overwhelming winner, with the uniform subsets law in second place. The greatest disparity between the estimates occurs for the file `bib`, which contains a formatted text bibliography. Here the best estimate (the uniform cardinality law) beats the worst estimate (Laplace's law) by 177 bytes. The uniform cardinality law looses by a significant amount on only one of the nineteen files (`pic`). The nonstationarity of that file causes problems for the strong predictions of the uniform cardinality law.

| file | size | $q$ | $nH\{n_i/n\}$ | $p_C()$ | $p_S()$ | $p_L()$ | $p_{\frac{1}{2}}()$ | A | B | C | D |
|---|---|---|---|---|---|---|---|---|---|---|---|
| bib | 111261 | 81 | 72330 | **92** | 102 | 269 | 174 | 111 | 170 | 152 | 121 |
| book1 | 768771 | 82 | 435043 | **116** | 127 | 352 | 219 | 118 | 188 | 181 | 133 |
| book2 | 610856 | 96 | 365952 | **124** | 132 | 329 | 212 | 138 | 220 | 191 | 156 |
| geo | 102400 | 256 | 72274 | 165 | 191 | 165 | **161** | 285 | 302 | 305 | 248 |
| news | 377109 | 98 | 244633 | **116** | 124 | 304 | 201 | 142 | 226 | 199 | 162 |
| obj1 | 21504 | 256 | 15989 | 129 | 147 | 129 | **126** | 249 | 194 | 150 | 174 |
| obj2 | 246814 | 256 | 193144 | 190 | 225 | 189 | **182** | 302 | 353 | 320 | 280 |
| paper1 | 53161 | 95 | 33113 | **100** | 108 | 236 | 156 | 118 | 161 | 141 | 117 |
| paper2 | 82199 | 91 | 47280 | **105** | 114 | 259 | 167 | 112 | 153 | 142 | 111 |
| paper3 | 46526 | 84 | 27132 | **92** | 101 | 238 | 154 | 103 | 142 | 130 | 103 |
| paper4 | 13286 | 80 | 7806 | **79** | 89 | 190 | 126 | 91 | 113 | 101 | 84 |
| paper5 | 11954 | 91 | 7376 | **83** | 89 | 181 | 122 | 102 | 124 | 104 | 92 |
| paper6 | 38105 | 93 | 23861 | **95** | 103 | 223 | 149 | 114 | 154 | 133 | 113 |
| pic | 513216 | 159 | 77636 | 216 | 194 | 323 | 205 | 170 | 171 | **102** | 131 |
| progc | 39611 | 92 | 25743 | **91** | 98 | 222 | 150 | 117 | 165 | 140 | 119 |
| progl | 71646 | 87 | 42720 | 97 | **85** | 253 | 164 | 110 | 158 | 114 | 112 |
| progp | 49379 | 89 | 30052 | **94** | 102 | 236 | 155 | 111 | 154 | 133 | 112 |
| trans | 93695 | 99 | 64800 | **105** | 113 | 252 | 169 | 130 | 190 | 166 | 137 |
| total: | 3251493 | — | 1786884 | **2089** | 2244 | 4350 | 2992 | 2623 | 3338 | 2904 | 2505 |

Table 1: Compression results on the Calgary corpus for eight parameter estimation rules, in whole bytes. All scores are relative to the empirical entropy $nH\{n_i/n\}$ of the file, which is a lower bound for these estimates. $q$ is the number of distinct symbols in each file. The uniform cardinality law $p_C()$ is overwhelmingly the most effective and the uniform subsets law $p_S()$ is the next most effective. Laplace's law $p_L()$ has by far the worst overall peformance. Of the four ad-hoc methods A-D, method D has the best average performance and method B has the worst performance.

The uniform cardinality law comes in second place to the uniform subsets law for the file `progl`, which contains a LISP program. The first 71 characters of this file are the LISP comment character (semi-colon), and so the uniform cardinality law predicts that the remainder of the file continue in this fashion. Unfortunately, this prediction only works until the actual text of the comment begins at the 75th character. Since the uniform subsets law converges slightly slower, it beats the uniform cardinality law on this file by a mere 12 bytes.

---
[8]The Calgary corpus is available via anonymous ftp from `ftp.cpsc.ucalgary.ca`.

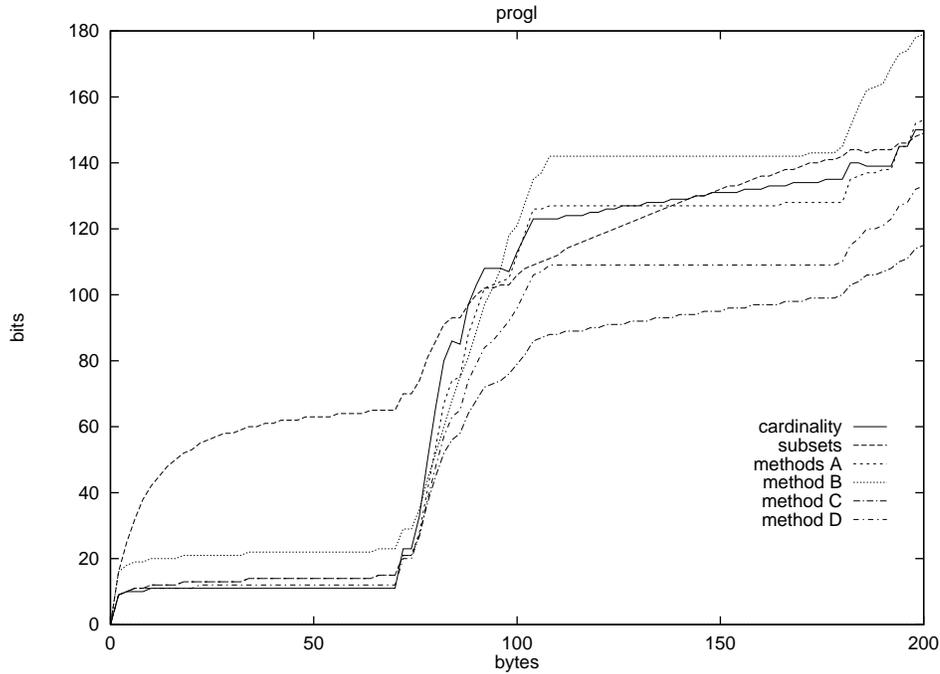

Figure 3: Compression rates for the first 100 bytes of the file `progl`, which contains a LISP program that begins with 71 semi-colons. Compressed bits are relative to the empirical entropy.

The uniform cardinality law comes in next to last place for the file `pic`, which contains a bit-mapped monochrome picture. Although this file contains 159 distinct characters, the first 52,422 bytes contain only 3 distinct characters. By the time the remaining 156 characters appear, our two laws of succession are extremely confident that the rest of the file will be limited to those first three characters. As a result, the uniform cardinality law looses to method C by 114 bytes on this file. This situation is best modeled by a small number states, and so should not be considered a weakness for either of our laws.

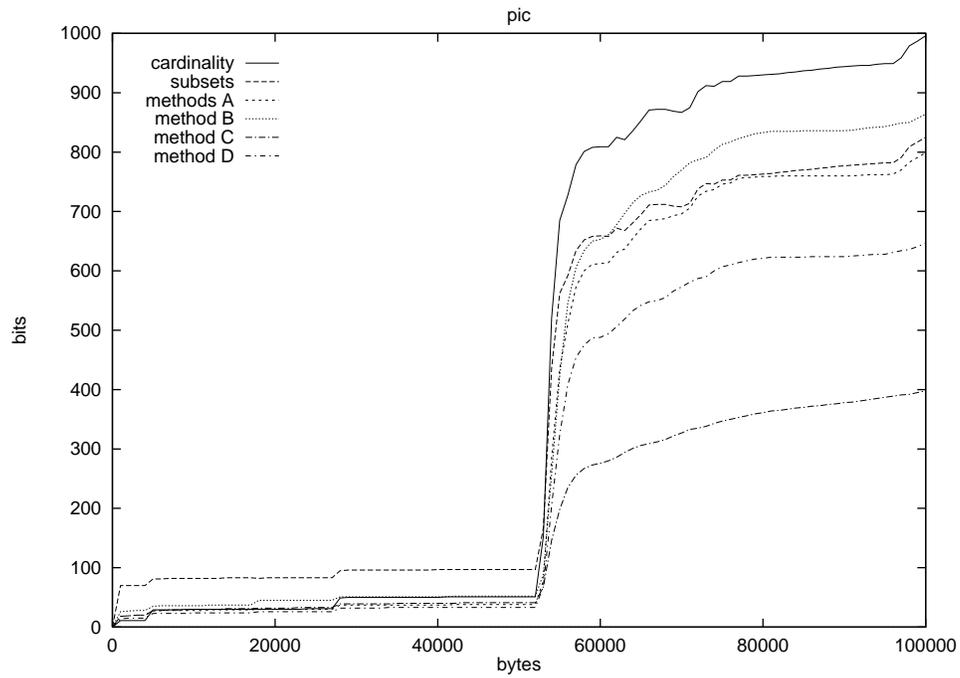

Figure 4: Compression rates for the first 100,000 bytes of the file `pic`, which contains a bit-mapped monochrome picture whose first 52,422 bytes contain only 3 distinct characters. Compressed bits are relative to the empirical entropy.

# 6  Conclusion

In this world, all strings are finite and most are very short. Alphabets are large. For this basic reason, natural strings do not include all the symbols in the alphabet. We have proposed a parameter-free natural law of succession (9) with this prior knowledge and proved that the total probability it assigns to the set of possible strings is within a constant factor of the probability assigned by any other probability function, and that the natural law outperforms Laplace's law by a factor of $\Theta((n + k - 1)^{k-q})$ for any finite string of length $n$. We have also shown that the natural law consistently provides the best predictions in real-world data compression applications.

# A  Good–Turing Estimates

I.J. Good and A. Turing [12] proposed the following estimate for enormous alphabets and many samples

$$p_{GT}(i|\{n_i\}, n) = \frac{n_i + 1}{n} \frac{f_{n_i+1}}{f_{n_i}} \tag{14}$$

where $f_j = |\{i : n_i = j\}|$ is the number of symbols that appeared exactly $j$ times. (Note that $f_0 = k - q$.) The principal motivation for the Good-Turing estimate is the desire to estimate the probablity of encountering an unseen symbol. The Good-Turing estimate performs poorly when the sample size is small, and so we do not consider it in the body of the paper.

When applied to multinomial estimation, the Good-Turing estimate also suffers from two defects. The first defect is that it will assign zero probability to frequent events when some of the $f_j$ are zero, as is likely to happen when $j$ is large. The second defect is that more frequent events may be assigned lower probabilities than less frequent events.

A popular ad-hoc solution to the first problem is to introduce a "discounting cutoff parameter," and to only perform Good-Turing discounting for those symbols whose frequencies fall below the value of the discounting cutoff parameter [22]. This approach will only work when the number of trials is significantly larger than the value of the cutoff parameter, and the $f_i$ values below the (value of the) cutoff parameter are stable. Consequently, it is not possible to use the Good-Turing estimate for online estimation or to assign probabilities to entire strings.

To remedy these problems, Ney and Essen [30] propose two variants of the Good-Turing estimate. In their *absolute discounting model*, the positive frequencies are decremented by a small constant amount $\delta$, $0 < \delta < 1$, and the remaining frequency is distributed uniformly across the novel events.

$$p_\delta(i|\{n_i\}, n) = \begin{cases} (n_i - \delta)/n & \text{if } n_i > 0 \\ q\delta/n(k-q) & \text{otherwise} \end{cases} \tag{15}$$

The total probability assigned to a string $x^n$, $x^n \in A^n$, by the absolute discounting models may be calculated directly

$$p_\delta(x^n|, n) = \frac{\delta^{q-1}(q-1)!(k-q)!}{k(k-1)!(n-1)!} \prod_{i \in q} \Gamma(n_i - \delta)/\Gamma(1 - \delta) \tag{16}$$

for a discounting constant $\delta$ independent of the string.

In the Ney-Essen *linear discounting model*, the positive frequencies are scaled by a small positive constant $\alpha$, $0 < \alpha < 1$, and the remaining frequency is again distributed uniformly across the novel events.

$$p_\alpha(i|\{n_i\}, n) = \begin{cases} (1-\alpha)n_i/n & \text{if } n_i > 0 \\ \alpha/(k-q) & \text{otherwise} \end{cases} \tag{17}$$

For constant $\alpha$, the total probability of a string is

$$p_\alpha(x^n|n) = \frac{\alpha^{q-1}(1-\alpha)^{n-q}(k-q)!}{k(k-1)!(n-1)!} \prod_{i \in q} \Gamma(n_i) \tag{18}$$

Note that the scaling value $\alpha$ must be $\Theta(1/n)$ for the linear discounting estimate to converge to the source probabilities at a reasonable rate. The immediate consequence of an adaptive $\alpha$ is that the total probability of a string will depend on the order of the symbols, and not just on their frequencies. A string whose symbols occur first early on will be more probable than a string whose symbols occur first later. Consequently, the symbol frequencies are not a sufficient statistic for this estimate when $\alpha \in \Theta(1/n)$.

The total probability assigned to novel events by these estimates is independent of the alphabet size $k$.

$$\begin{array}{rcl} p_{GT}(\bar{q}|\{n_i\}, n) & = & f_1/n \\ p_\delta(\bar{q}|\{n_i\}, n) & = & q\delta/n \\ p_{\hat\alpha}(\bar{q}|\{n_i\}, n) & = & \alpha \end{array} \tag{19}$$

Let us first consider the ratio of the probabilities of the absolute and linear discounting models for fixed $\delta$ and $\alpha$.

$$\begin{array}{rcl} p_\delta(x^n|n)/p_\alpha(x^n|n) & = & \dfrac{\delta^{q-1}(q-1)!}{\alpha^{q-1}(1-\alpha)^{n-q}\Gamma(1-\delta)^q} \prod_{i \in q} \Gamma(n_i - \delta)/\Gamma(n_i) \\ & = & \Theta(((1-\alpha)^{n-q} \prod_{i \in q}(n_i - \delta)^\delta)^{-1}) \\ & > & \Theta(((1-\alpha)^{n-q}(n - q\delta)^{q\delta})^{-1}) \\ & = & \Theta(2^{n-q}) \end{array}$$

This analysis makes clear that the linear discounting model for fixed $\alpha$ is simply not a viable estimate.

So it is more interesting to compare the absolute discounting model to the uniform cardinality law.

$$\begin{array}{rcl} p_\delta(x^n|n)/p_C(x^n|n) & = & \dfrac{\delta^{q-1}k!(n-1)!n!}{q!(k-q)!(q-1)!(n-q)!} \prod_{i \in q} \Gamma(n_i - \delta)/\Gamma(n_i + 1) \\ & = & \Theta((n-q)^q / \prod_{i \in q} n_i^{1+\delta}) \end{array}$$

We note that

$$n^{1+\delta} \leq \prod_{i \in q} n_i^{1+\delta} < n^{q(1+\delta)}$$

with the immediate implication that

$$\Theta(1/n^{q\delta}) \leq p_\delta(x^n|n)/p_C(x^n|n) \leq \Theta(n^{q-1-\delta}).$$

Thus, the relative performance of the absolute discounting model and uniform cardinality law is a function of the entropy $H(\{n_i/n\})$ of the relative frequencies. When the empirical entropy is high, then the uniform cardinality law will do better; when the empirical entropy is low, then the absolute discounting model will do better.